# Gaze Estimation on Spresense

Thomas Rüegg
D-ITET, ETH Zurich
Zurich, Switzerland
rueeggth@ethz.ch

Pietro Bonazzi
D-ITET, ETH Zurich
Zurich, Switzerland
pbonazzi@ethz.ch

Andrea Ronco
D-ITET, ETH Zurich
Zurich, Switzerland
aronco@ethz.ch

*Abstract*—Gaze estimation is a valuable technology with numerous applications in fields such as human-computer interaction, virtual reality, and medicine. This report presents the implementation of a gaze estimation system using the Sony Spresense microcontroller board and explores its performance in latency, MAC/cycle, and power consumption. The report also provides insights into the system's architecture, including the gaze estimation model used. Additionally, a demonstration of the system is presented, showcasing its functionality and performance. Our lightweight model TinyTrackerS is a mere 169Kb in size, using 85.8k parameters and runs on the Spresense platform at 3 FPS.

## I. Introduction

Gaze estimation is the process of determining where a person is looking. By accurately estimating the gaze direction, it becomes possible to understand user intentions, enhance human-computer interaction, and develop immersive experiences. In this report, we explore the implementation of a gaze estimation system using the Sony Spresense microcontroller board, which provides a compact and power-efficient platform for deploying such systems.

## II. Architecture

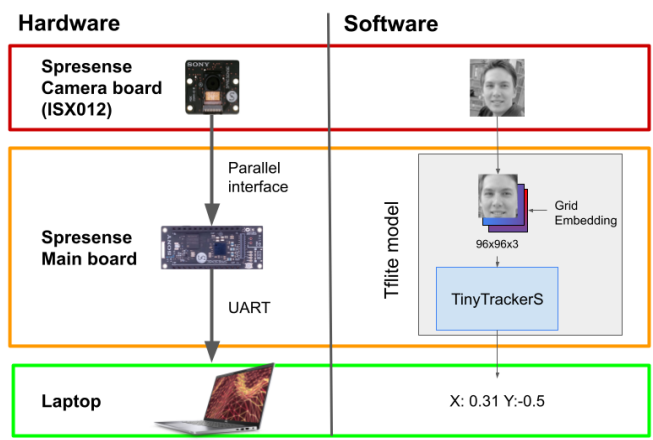

Fig. 1. Overview

### A. Network

The gaze estimation model used in this system is a lightweight deep-learning model based on [1] optimized for deployment on the Spresense Microcontroller. In Tab. I, we compare our model to its predecessors in terms of input resolution, model weight footprint, MAC (Multiply-Accumulate), and number of parameters.

Our model, TinyTrackerS, features a fully convolutional backbone, combining both standard and depthwise convolutional layers, followed by a dense detection network. Compared to its predecessor, we drastically reduce layers and simplify the backbone. As input the model uses Greyscale images (96x96) with grid embeddings as detailed in [1]. During deployment instead of using a dynamic face detection algorithm to find and crop the face, we use a constant crop in the center of the image resulting in a fixed grid embedding that is stored within the tflite model. Therefore for the demo, only a greyscale image is required as input.

The model predicts gaze locations directly from greyscale images. The outputs are clamped to -10 to 10 cm and normalized to [-1, 1]. This results in improved resolution of the prediction after quantization. The model is trained on the GazeCapture dataset [2] using random augmentations and early stopping to avoid overfitting and improve precision. Finally, the model is converted to tflite and quantized to 8-bit integers, retaining full precision only at the output layer.

TABLE I
Model comparison

| Name | Input res. | Params | MAC | Size[KB] |
|---|---|---|---|---|
| iTracker [2] | 224x224 | 6'287k | 2651M | 24'600 |
| TinyTracker [1] | 112x112 | 455k | 11.8 M | 608 |
| TinyTrackerS | 96x96 | 85.8k | 5.3 M | 169 |

### B. Implementation Details

The gaze estimation system has been implemented on the Sony Spresense microcontroller board combined with the ISX012 camera board, which offers a powerful yet energy-efficient platform for edge computing applications. The system utilizes the Sony SDK and TensorFlow Lite framework for efficient model execution. The datasets used for training and evaluation have been carefully selected to ensure robust performance across different individuals and scenarios.

The camera board is connected to the main board via a ribbon cable [3], allowing direct fast parallel transfers of images.

During the image capture process the image is cropped, resized, and loaded directly into the input tensor of the tflite model. The subsequent inference process is accelerated using optimized CMSIS NN operations.

In total the demo uses 463.5 Kb of FLASH memory, allocating 128 Kb to tensors during model inference.

## III. RESULTS

The performance of the gaze estimation system was evaluated in terms of latency, MAC/cycle, power efficiency, and model accuracy. We report metrics in terms of end to end and inference only performance. The latency measures the delay between input acquisition and output generation, while MAC/cycle indicates the computational efficiency of the system. Power efficiency quantifies the energy consumption of the system during operation. Results and comparison to [1] can be found in Tab. II. Model accuracy is assessed following the same procedure as in [1], reporting the precision of the model as the average prediction error in centimeters reported in Tab. III and comparing it to [1].

We find that our TinyTrackerS model reduces latency and energy consumption of the inference process by 1.9x. End to end we achieve a reduction of 1.5x. However, inference Efficiency is slightly decreased.

This gain in performance comes at a trade-of in precision, where we report a drop of 7% compared to TinyTracker.

TABLE II
HARDWARE EVALUATION ON SPRESENSE

| Platform | TinyTracker [1] | TinyTrackerS |
|---|---|---|
| End-to-End Evaluation | | |
| E [mJ] ↓ | 234.1 | 146.5 |
| Latency [ms] ↓ | 522.5 | 327 |
| Inference Evaluation | | |
| MAC/Cycle ↑ | 0.20 | 0.17 |
| Latency [ms] ↓ | 386.60 | 197 |
| E [mJ] ↓ | 31.97 | 16.29 |

TABLE III
MODEL PRECISION

| | iTracker [2] | TinyTracker [1] | TinyTrackerS |
|---|---|---|---|
| Error [cm] | 2.46 | 2.62 | 2.82 |

## IV. DEMO

For the demo, we 3D printed a case for the Spresense Main Board and Camera [4]. The material we used was a white PLA V and was printed with a Prusa i3 MK3S. The case and the Spresense board were the connected to a computer. In the computer screen, the user is able to visualize the gaze predictions. The entire image capture and neural network inference processing is done on the Spresense board and are then trasmitted over UART to the computer.

In Table IV we show qualitative data, related to our experimental evaluation. We saved the images and prediction for a user looking at different regions of the screen. We overlaid a blue-dot on the images to visualize the prediction of the network.

TABLE IV
QUALITATIVE RESULTS

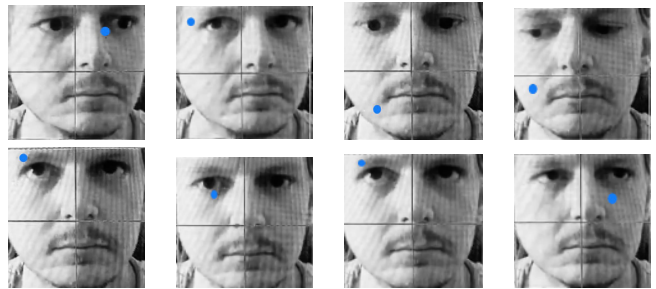

TABLE V
3D PRINTED CASE [4]

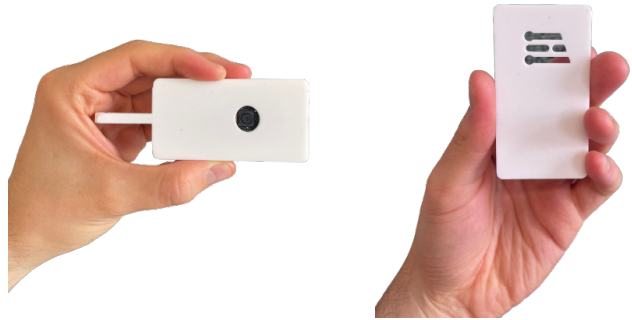

## V. CONCLUSION

In this report, we provided a demo on how Sony Spresense can be used to estimate 2D gaze. Our proposed model significantly improves upon it's predecessors in terms of inference speed running at 3 frames per second, while trading of 7% in precision.